\documentclass{article}
\usepackage{spconf,amsmath,graphicx}
\usepackage{multirow}
\usepackage{amsfonts}
\usepackage{float}
\usepackage{subcaption}
\usepackage{algorithm}
\usepackage{algorithmic}

\usepackage{enumitem}
\setlist{nosep, leftmargin=14pt}

\usepackage{mwe} 

\def\L{{\cal L}}
\def\ie{{i.e.~}}
\def\al{et al.~}
\def\eg{e.g.,~}
\usepackage{commands}
\title{ NC-Reg : Neural Cortical Maps for Rigid Registration} 

\name{\begin{minipage}{0.7\linewidth}
    \centering Ines Vati$^{1,2}$, Pierrick Bourgeat$^1$, Rodrigo Santa Cruz$^2$, Vincent Dore$^1$, Olivier Salvado$^2$, Clinton Fookes$^2$, Léo Lebrat$^2$
\end{minipage}}
\address{$^1$ CSIRO, Australia\hspace{3cm} $^2$Queensland University of Technology, Australia }
\begin{document}
%
\maketitle
\begin{abstract}
We introduce \textit{neural cortical maps}, a continuous and compact neural representation for cortical feature maps, as an alternative to traditional discrete structures such as grids and meshes. It can learn from meshes of arbitrary size and provide learnt features at any resolution. \textit{Neural cortical maps} enable efficient optimization on the sphere and achieve runtimes up to 30 times faster than classic barycentric interpolation (for the same number of iterations). As a proof of concept, we investigate rigid registration of cortical surfaces and propose NC-Reg, a novel iterative algorithm that involves the use of \textit{neural cortical feature maps}, gradient descent optimization and a \textit{simulated annealing} strategy. Through ablation studies and subject-to-template experiments, our method demonstrates sub-degree accuracy ($<1^\circ$ from the global optimum), and serves as a promising robust pre-alignment strategy, which is critical in clinical settings. 
\end{abstract}
\section{Introduction and Related Work}
\label{sec:intro}


Accurate cortical surface Non-Rigid Registration (NRR) is crucial for comparing brain anatomy across subjects, enabling population studies and clinical applications. The main drawbacks of existing NRR methods are the intricate processing of spherical meshes, commonly used to model the brain cortex \cite{fischl_freesurfer_2012}, and a strong reliance on an accurate initial rigid alignment.

For example, Spherical Demons (SD) \cite{yeo_spherical_2010}, which achieves competitive runtimes compared to FreeSurfer (FS) \cite{fischl_freesurfer_2012} (which takes approximately 30 minutes for spherical registration \cite{suliman_geomorph_2022, ren_sugar_2024}), performs rotation searches at each multi-resolution level. 
Hierarchical Spherical Deformation (HSD) \cite{lyu_hierarchical_2019} jointly optimizes global rigid alignment and local deformation using spherical harmonics, ensuring smooth transformations. Those multi-resolution, iterative approaches achieve high accuracy but remain slow.
Even faster deep learning-based methods still rely on error-prone remeshing and rigid pre-alignment. For instance, S3Reg \cite{zhao_s3reg_2021} requires substantial computational resources (especially during training) to enable their proposed convolution and pooling operations on meshes, and performs rigid registration via exhaustive angle searches similarly to SD and FS.
Other methods, such as DDR \cite{suliman_deep-discrete_2022}, GeoMorph \cite{suliman_geomorph_2022} and SUGAR \cite{ren_sugar_2024}, integrate rotation prediction within their models, but they often lack generalization when spherical meshes are largely misaligned.

As previously noted, most neuroimaging pipelines, including FS \cite{fischl_freesurfer_2012}, ANTs \cite{tustison_large-scale_2014}, and BrainSuite \cite{shattuck_brainsuite_2002}, represent the cortical surface as discrete triangular meshes composed of vertices, edges, and faces. Typical cortical meshes contain about 140,000 vertices per hemisphere.
To reduce computational complexity, most approaches downsample these high-resolution surfaces to regular icosahedral meshes \cite{fawaz_benchmarking_2021}, then upsample the final deformation using Thin Plate Spline interpolation (\eg MSM \cite{robinson_msm_2014}) or linear interpolation (\eg SD \cite{yeo_spherical_2010}). Classical optimization on discrete meshes requires barycentric interpolation at each iteration to warp the moving surface, which is computationally intensive and error-prone. A common workaround is to project surface features onto 2D images for interpolation, but this introduces distortions. Large surface regions can map to sub-pixel areas \cite{morreale_neural_2021}, ultimately degrading accuracy.
To address these limitations, learning-based representations have emerged in computer vision as continuous and compact alternatives to discrete structures.
Yang \al \cite{yang_geometry_2023} explored geometric processing on \textit{neural implicit fields}, deep networks predicting signed distance functions.
\textit{Overfitted} neural networks (NN), \ie trained on a single surface, have also been proposed \cite{morreale_neural_2021}, leveraging the universal approximation theorem \cite{hornik_multilayer_1989}. More recently, Williamson \al \cite{williamson_neural_2025} introduced geometric processing tools operating directly on Spherical Neural Surfaces (SNS), exploiting the smooth, differentiable nature of neural networks to model genus-0 surfaces without remeshing. DeepCSR \cite{cruz_deepcsr_2021} predicted implicit surface representations of the brain cortex, but \textit{overfitted} neural models remain under-explored in neuroimaging.

Inspired by neural radiance fields (NeRF) \cite{mildenhall_nerf_2020}, we propose \textit{Neural Cortical maps} (NC), a continuous and compact representation for modeling cortical feature maps. Our NN takes coordinates on the sphere as input and outputs features of interest. Once trained on a mesh of arbitrary size, the model can provide features at any cortical location without mesh interpolation. 
To demonstrate its practical utility, we propose NC-Reg, a novel rotational registration method using NC, and a \textit{Simulated Annealing} (SA) strategy \cite{kirkpatrick_optimization_1983}. We show that NC-Reg is efficient, fast and accurate. It is also a robust pre-alignment strategy that enhances downstream NRR performance. Our contributions are (1) a novel neural representation for cortical maps that trains easily on meshes of any size and generates features on meshes at arbitrary resolutions, (2) a fast iterative algorithm for global rigid registration on the sphere, and (3) a robust initialization strategy for NRR.


\begin{figure}[htb]
\begin{minipage}[b]{\linewidth}
    \centering
    \includegraphics[height=.2\textheight]{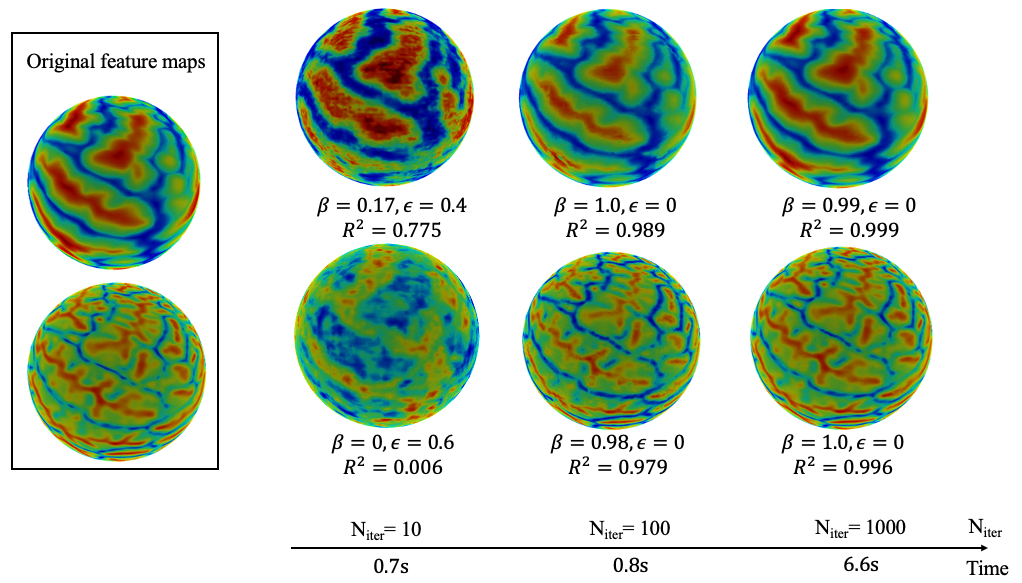}
\end{minipage}
    \caption{Learning two cortical feature maps using different numbers of training iterations ($N_\text{iter}$). To evaluate cortical feature reconstruction, we realised a linear regression between the interpolated values and the model outputs for an icosahedron of level 6 (40,962 vertices). $\beta$ is the slope, $\epsilon$ the intercept, and $R^2$ the coefficient of determination. The NN recovers both large-scale geometric features (average convexity, top) and high-frequency features (mean curvature, bottom).}
    \label{fig:overfit_niter}
\end{figure}

\section{Method}
\label{sec:method}

\subsection{Definition and Training of \textit{Neural Cortical Maps}}
We are interested in learning a continuous representation of cortical surfaces.
We define a Multi-Layer Perceptron (MLP) parametrized by $\varphi$, referred to as a \textit{neural cortical map} (NC), $\s_\varphi: \sphere^2 \rightarrow \reals^{n_\textbf{f}}$, where $\sphere^2$ is the $2$-dimensional unit sphere and $n_\textbf{f}$ is an arbitrary number of features (\eg mean curvature, convexity, sulcal depth and/or cortical thickness).
The input $p$ of this neural network is a point lying on the unit sphere $p \in\sphere^2$. 
The input is further encoded using a multiresolution hash encoding \cite{muller_instant_2022}. The deep learning model outputs the feature values at $p$. In practice, its architecture is the fast fully-fused MLP\footnote{https://github.com/NVlabs/tiny-cuda-nn} from \cite{muller_real-time_2021}.

Let $\G = (\V, \F)$ be the original spherical mesh where $\V$ is the set of vertices and $\F$ is the set of faces. The map $\mathbf{f}:\V\rightarrow \reals^{n_\textbf{f}}$ gives the feature values for a given vertex.
At each training step, we sample $n_{\F}$ faces on the original mesh $\G$ and $n_{\V}$ random points on each sampled face. Let $p_1, p_2, p_3\in\V$ be the vertices of the sampled face. A sampled point is obtained by $p = ap_1 + bp_2 + cp_3$ where $(a,b,c)$ are random scalars in the standard 2-simplex $\Delta^2$. 
The target feature values of a random point $p$ is obtained via barycentric interpolation  
\begin{equation}\label{eq:interp}
 I(p) = \frac{|p_2p_3p|\mathbf{f}(p_1) +|p_1p_3p|\mathbf{f}(p_2) +|p_1p_2p|\mathbf{f}(p_3)}{|p_2p_3p| + |p_1p_3p|+ |p_1p_2p|}
\end{equation}
where $|p_ip_jp|$ is the area of the triangle $[p_i, p_j, p]$.


A gradient descent approach is performed to optimize the network parameters $\varphi$ using Adam optimizer \cite{kingma_adam_2014} that minimizes a feature matching energy such as the Mean Squared Error (MSE) between $S_\varphi$ and $I$.

\figref{fig:overfit_niter} shows some results of the training on a given subject. The representation can also be seen as a compression for large meshes. For instance, a mesh of $136$,$804$ vertices has a size of $4.57$ MB and each feature file is $0.51$ MB. The MLP would have about $3.7$ MB (adding more output features changes only the last layer, which is negligible in terms of memory), while allowing for fast reconstruction of a subject's cortical mesh for any resolution. 


\subsection{Global Rotational Registration Algorithm}

Let $F:\sphere^2\rightarrow\reals^{n_\textbf{f}}$ be the fixed feature map and $M:\sphere^2\rightarrow\reals^{n_\textbf{f}}$ be the moving feature map.
We aim at finding the optimal rotation $R\in SO(3)$ where $SO(3)$ is the Special Orthogonal group such that $RR^T=R^TR=I$ and $\text{det}(R)=1$ that solves
$\umin{R\in SO(3)}{\int_{\sphere^2}\L(F(p), M\circ R(p))dp}$. This integral is approximated by sampling $N$ points on the sphere. The energy $\L_t$ at iteration $t$ for a rotation $R_t$ reads
\begin{equation}\label{eq:rot_disc_opt}
\L(F, M, R_t)=\frac{1}{N}\sum_{i=1}^N\text{MSE}\big(F(p_i)-M( R_t\cdot p_i)\big)
\end{equation}
Minimizing \eqnref{eq:rot_disc_opt} is non convex and optimizing rotations is usually problematic, mainly because they are non-Euclidean in nature \cite{grassia_practical_1998}.
To handle the fact that many local minima exist, we propose a random exploration of the rotation space, \ie every time the algorithm is "stuck" within a local minimum, a random resetting of the rotation parameters is proposed and the gradient descent starts from this new initialization. The resetting is either completely random (w/o SA) or subject to a certain acceptance probability (SA) as described in \algref{alg:anneal}.

Many rotation parametrizations exist. One can mention Euler angles, quaternions, and axis-angle parametrization. Zhou \al \cite{zhou_continuity_2019} also introduced the continuous 6D representation for 3D rotations estimation. 
The rotation model is "reset" by converting a random axis-angle representation. The rotation angle is uniformly sampled in $[0,2\pi]$ and the rotation axis is obtained by normalizing a Gaussian vector. 

The moving average $\bar{\L}_t = \sum_{i=t-t_m}^{t-1}\L_i$ of the loss function \eqref{eq:rot_disc_opt} is computed at each iteration $t$. A local minima is identified when the difference $|\L_t-\bar{\L}_t|$ is below a certain threshold for a certain number of iterations $t_{\text{diff}}$. We keep the local minimum with the lowest validation loss.

\begin{algorithm}
\caption{SA reset strategy. A temperature parameter $T_t$ is decreased every time a local minimum is identified until a temperature minimum $T_{\min}$ is reached. The idea is that when close to the end of the optimization, the probability of accepting a worse candidate is smaller. In practice, $T_{\min}=10^{-4}$, $\alpha_T=0.9$ ($0<\alpha_T<1$), $N_{T, \text{iter}}=5$.}
\label{alg:anneal}
    \begin{algorithmic}[1]
    \FOR{$k\in\intinter{1}{N_{T, \text{iter}}}$}
    \STATE  $R_{\text{rand}}\sim \U(SO(3))$ 
    \STATE $\Delta\L = \L(\brackidx{p_i}{1}{N}, F, M, R_{\text{rand}}) - \L_t$
    \IF{$\Delta\L < 0$ \textbf{OR} $\left(\Delta\L > 0 \text{ and } \U([0,1]) < \exp(\frac{-\Delta\L}{T_t})\right)$}
        \STATE $R_{t+1}\gets R_\text{rand}$
        \STATE \textbf{break}
    \ENDIF
    \ENDFOR
    \IF{$T_t > T_{\min}$}
        \STATE $T_{t+1} \gets \alpha_TT_{t}$
    \ENDIF
    \end{algorithmic}
\end{algorithm}


\section{Experiments and Results}
\label{sec:results}
In \figref{fig:overfit_niter}, after \textit{overfitting} a subject from the ADNI dataset, a linear regression between model outputs and interpolated values of new vertices yields a near-identity function, confirming the fidelity of the learned representation. In the following, we focus on the spherical registration problem. We trained a NC model on the \textit{fsaverage} template from FreeSurfer v7.4 over $3$,$000$ iterations for about 18s.

We use the Washington University Alzheimer’s Disease Research Center (ADRC) dataset, which includes $39$ cortical meshes with expert manual parcellations and compare the rigid and non-rigid alignment of each subject to the template. To assess robustness to initialization, we also created a \textit{perturbed} dataset by applying random rotations to each cortical mesh: (roll, yaw, pitch) $\sim\U([-36^\circ, 36^\circ]^3)$).
To evaluate the alignment quality, we compute the MSE and the Pearson Correlation Coefficient (PCC) for two cortical features: mean curvature (curv) and convexity (sulc). We also compute the Dice score of the template and registered subject parcellations. 
To compare two rotations $R_1, R_2\in SO(3)$, we convert them to their quaternion representation ($q_1,q_2\in \{q\in\mathbb{R}^4, ||q||^2=1\}$) and compute the quaternion error as it is bi-invariant and respects the topology on $SO(3)$ \cite{huynh_metrics_2009}
\begin{equation}\label{eq:dist_R}
    \text{dist}_R(q_1,q_2)= \text{arccos}(|q_1\cdot q_2|)\in[0,\pi/2]\subset\mathbb{R}^+
\end{equation}
 Experiments were performed on a NVIDIA H100 GPU, except for SD which was run in Matlab R2024b using fine-tuned parameters.
 
\subsection{Rigid Registration - Ablation Study}
\begin{figure}[htb]
\centering
\begin{minipage}[b]{\linewidth}
\includegraphics[width=\linewidth]{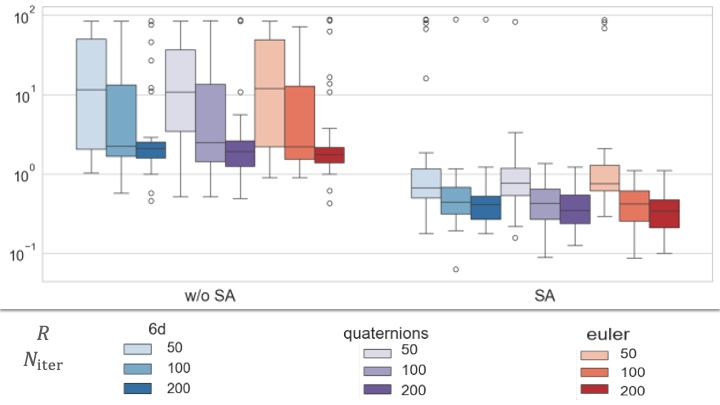}
\end{minipage}
\caption{Rotation distance $\text{dist}_R(R^*, R^*_{perturbed}\cdot R_{\text{rand}})$ in degrees (\eqnref{eq:dist_R}) between the solutions obtained on the same subject for two different initializations. For each subject, $R^*$ is the optimized solution (for the default initialization). $R_{\text{rand}}$ is the random perturbation applied to the subject mesh. $R^*_{perturbed}$ is the optimized solution to register the \textit{perturbed} subject to the template.}
\label{fig:sanity_reg}
\end{figure}

As expected, increasing the number of iterations increases the chance of converging to the global optimum (\figref{fig:sanity_reg}). 
Our algorithm shows strong robustness and consistency. With $N_{\text{iter}}\geq100$, solutions from different initializations remain within $1^\circ$ of each other. 
Although SA adds less than one second to runtime, it substantially improves feature alignment and ensures successful registration across all subjects (\tabref{tab:ablation_reset}).
The choice of rotation model showed no significant impact. In the following experiments, we use SA, quaternion model and $N_{\text{iter}}=100$.

\begin{table}[htb]
    \caption{Comparison of \textit{reset} strategies for NC-Reg on the \textit{perturbed} dataset. "Worst Dice" indicates the proportion of subjects for which the Dice score decreased after registration. Results use $N_{\text{iter}}=200$ and quaternion parametrization as it yielded the best performance across all methods. Two-tailed paired-sample $t$-tests compare MSE, PCC and Dice values between SA and other strategies (*$p<0.005$), with 95\% confidence intervals shown.}
	\label{tab:ablation_reset}
	\begin{center}
    \resizebox{\columnwidth}{!}{
		\begin{tabular}{|c|c|c|c|c|c|}
			\hline
			\rule{0pt}{3ex}\Large{Method} & \Large{MSE $(\downarrow)$} & \Large{PCC $(\uparrow)$} & \Large{Dice $(\uparrow)$}& \Large{Time (s)} & \Large{Worst Dice} \\
			\hline\hline
            \multicolumn{6}{|c|}{\rule{0pt}{3ex} \Large{\textit{perturbed} dataset}} \\[0.5ex]
            \hline\hline
            \rule{0pt}{3ex}\Large{w/o Reset} & $\scalebox{1.5}{0.156} \pm 0.030$* & $\scalebox{1.5}{0.403} \pm 0.067$* & $\scalebox{1.5}{0.559} \pm 0.096$* & $\scalebox{1.5}{0.63} \pm 0.002$ & $\scalebox{1.5}{6/ 39 (15.4\%)}$ \\\hline
			\rule{0pt}{3ex}{\Large w/o SA} & $\scalebox{1.5}{0.125} \pm 0.015$* & $\scalebox{1.5}{0.454} \pm 0.037$* & $\scalebox{1.5}{0.677} \pm 0.055$* & $\scalebox{1.5}{0.69} \pm 0.008$ & $\scalebox{1.5}{2/ 39 (5.1\%)}$ \\
            \rule{0pt}{3ex}\Large{SA} & $\scalebox{1.5}{\textbf{0.098}} \pm 0.007$ & $\scalebox{1.5}{\textbf{0.532}} \pm 0.021$ & $\scalebox{1.5}{\textbf{0.761}} \pm 0.019$ & $\scalebox{1.5}{1.48} \pm 0.01$ & $\scalebox{1.5}{0/ 39 \textbf{(0.0\%)}}$ \\
			\hline
		\end{tabular}
    }
	\end{center}	
\end{table}

\subsection{Robust Rigid Pre-Alignment}
\begin{table*}[hbt]
    \caption{Comparison to registration baselines. Total time for SUGAR includes the time for interpolating the given subject mesh to the icosahedron of level $6$ (before inputting it to the network) and the inference time. We performed two-tailed paired-sample $t$-tests between each rigid method and NC-Reg (*$p<0.001$), between SD and NC-Reg+SD (• $p<0.005$), and between SUGAR and NC-Reg+SUGAR ($^\dagger p<0.005$).}
	\label{tab:comp_baselines}
	\begin{center}
    \resizebox{2\columnwidth}{!}{
		\begin{tabular}{|c|c|c|c|c|c|c|c|}
			\hline
			\multirow{2}{*}{Method} & \multicolumn{2}{c|}{MSE $(\downarrow)$}  & \multicolumn{2}{c|}{PCC $(\uparrow)$}  & \multirow{2}{*}{Dice $(\uparrow)$} & \multirow{2}{*}{Time (s)} & \multirow{2}{*}{Worst Dice} \\
			\cline{2-5}
			 & curv & sulc & curv & sulc &  &  &  \\
			\hline\hline
            \multicolumn{8}{|c|}{\rule{0pt}{2ex} Rigid registration on the ADRC dataset} \\[0.5ex]
            \hline\hline
			NC-Reg (ours) & $0.027 \pm 0.0008$ & $0.169\pm 0.014$ & $0.328 \pm 0.022$ & $0.735\pm 0.021$ & $0.761 \pm 0.018$ & $0.84 \pm 0.01$ & 0.0\% \\
            Interp-Reg & $0.027 \pm 0.0008$ & $0.169\pm 0.014$ & $0.329 \pm 0.0215$ & $0.735\pm 0.021$ & $0.760 \pm 0.018$ & $48.67 \pm 0.48$ & 0.0\% \\
            \hline
            Rigid SUGAR & $0.028 \pm 0.0010$* & $0.181\pm 0.016$* & $0.297 \pm 0.0255$* & $0.716\pm 0.024$* & $0.760 \pm 0.014$ & $1.55 \pm 0.07$ & 0.0\% \\
			\hline
            Rigid HSD (ico4) & $0.027 \pm 0.0008$ & $0.168\pm 0.013$* & $0.327 \pm 0.0213$ & $0.738\pm 0.021$* & $0.760 \pm 0.017$ & $5.98 \pm 0.08$ & 0.0\% \\
            Rigid HSD (ico6) & $0.027 \pm 0.0009$ & $0.167\pm 0.013$* & $0.329 \pm 0.0218$ & $0.738\pm 0.021$* & $0.761 \pm 0.017$ & $14.85 \pm 0.39$ & 0.0\% \\
			\hline\hline
            \multicolumn{8}{|c|}{\rule{0pt}{2ex} Rigid registration on the \textit{perturbed} dataset} \\[0.5ex]
            \hline\hline
            NC-Reg (ours) & $0.027 \pm 0.0008$ & $0.170\pm 0.014$ & $0.327 \pm 0.022$ & $0.734\pm 0.021$ & $0.760 \pm 0.018$ & $0.79 \pm 0.01$ & 0.0\% \\
			Interp-Reg & $0.028 \pm 0.0011$ & $0.193\pm 0.030$ & $0.306 \pm 0.0288$ & $0.697\pm 0.048$ & $0.715 \pm 0.054$ & $44.59 \pm 1.59$ & 5.1\% \\
			\hline
            Rigid SUGAR & $0.038 \pm 0.0010$* & $0.508\pm 0.046$* & $0.043 \pm 0.0247$* & $0.200\pm 0.072$* & $0.306 \pm 0.066$* & $1.24 \pm 0.05$ & \underline{12.8\%} \\
			\hline
            Rigid HSD (ico4) & $0.027 \pm 0.0009$ & $0.167\pm 0.013$* & $0.328 \pm 0.0218$ & $0.738\pm 0.021$* & $0.760 \pm 0.017$ & $8.59 \pm 0.26$ & 0.0\% \\
			\hline\hline
            \multicolumn{8}{|c|}{\rule{0pt}{2ex} Non-rigid registration on the ADRC dataset} \\[0.5ex]
            \hline\hline
            SD & $0.017 \pm 0.0004$ & $0.056\pm 0.003$ & $0.580 \pm 0.009$ & $0.914\pm 0.004$ & $0.820 \pm 0.007$ & $37.93 \pm 0.47$ & 0.0\% \\
            NC-Reg + SD & $0.017 \pm 0.0004$ & $0.057\pm 0.004$ & $0.581 \pm 0.0098$ & $0.913\pm 0.006$ & $0.819 \pm 0.008$ & $31.70 \pm 0.44$ & 0.0\% \\
			\hline
            SUGAR & $0.018 \pm 0.0004$ & $0.062\pm 0.002$ & $0.546 \pm 0.008$ & $0.905\pm 0.003$ & $0.825 \pm 0.006$ & $1.79 \pm 0.07$ & 0.0\% \\
            NC-Reg + SUGAR & $0.018 \pm 0.0004$ & $0.063\pm 0.003$ & $0.545 \pm 0.0090$ & $0.903\pm 0.005$ & $0.822 \pm 0.009$ & $2.40 \pm 0.04$ & 0.0\% \\
			\hline
            HSD & $0.019 \pm 0.0004$ & $0.066\pm 0.003$ & $0.534 \pm 0.008$ & $0.900\pm 0.004$ & $0.811 \pm 0.007$ & $17.28 \pm 0.27$ & 0.0\% \\
			\hline
			\hline\hline
            \multicolumn{8}{|c|}{\rule{0pt}{2ex} Non-rigid registration on the \textit{perturbed} dataset} \\[0.5ex]
            \hline\hline
            SD & $0.017 \pm 0.0004$ & $0.057\pm 0.003$ & $0.574 \pm 0.010$ & $0.912\pm 0.005$ & $0.812 \pm 0.008$ & $35.55 \pm 0.26$ & 0.0\% \\
			NC-Reg + SD & $0.017 \pm 0.0004$• & $0.056\pm 0.003$ & $0.582 \pm 0.0091$• & $0.914\pm 0.004$ & $0.821 \pm 0.007$• & $30.26 \pm 0.35$ & 0.0\% \\
			\hline
            SUGAR & $0.025 \pm 0.0012$ & $0.167\pm 0.025$ & $0.388 \pm 0.031$ & $0.731\pm 0.042$ & $0.412 \pm 0.089$ & $1.46 \pm 0.05$ & \underline{7.7\%} \\
            NC-Reg + SUGAR & $\textbf{0.018} \pm 0.0004$$^\dagger$ & $\textbf{0.063}\pm 0.003$$^\dagger$ & $\textbf{0.545} \pm 0.0089$$^\dagger$ & $\textbf{0.903}\pm 0.005$$^\dagger$ & $\textbf{0.822} \pm 0.009$$^\dagger$ & $2.13 \pm 0.05$ & 0.0\% \\
			\hline
            HSD & $0.019 \pm 0.0004$ & $0.066\pm 0.003$ & $0.534 \pm 0.008$ & $0.900\pm 0.004$ & $0.811 \pm 0.007$ & $17.27 \pm 0.30$ & 0.0\% \\
			\hline
		\end{tabular}
    }
	\end{center}	
\end{table*}

We benchmarked our rigid registration method against two state-of-the-art approaches: the pre-trained SUGAR model \cite{ren_sugar_2024} and the CUDA implementation of HSD algorithm \cite{lyu_hierarchical_2019} for the first degree of freedom ($l=0$, see Section 2.3 in \cite{lyu_hierarchical_2018}). 

\tabref{tab:comp_baselines} shows that NC-Reg achieves comparable or superior performance to state-of-the-art methods. It significantly outperforms SUGAR on the \textit{perturbed} datasets as deep-learning models usually do not generalize well. Rigid HSD runtime highly depends on the chosen resolution, while our approach is consistently faster. 

For NRR, we replace the rigid component of SD and SUGAR with our rotational pre-alignment.
We observed that replacing the rotation search in SD by the initialization obtained with NC-Reg decreases the time by about 5s and yields better Dice score for the \textit{perturbed} dataset.
Using NC-Reg pre-alignment drastically improves SUGAR results on the \textit{perturbed} dataset. 
Overall, NC-Reg provides an efficient and reliable initialization, improving both rigid and non-rigid registration pipelines.

\section{Conclusion and Discussion}
\label{sec:ccl}

We introduced NC maps, a new continuous representation for cortical feature maps, and demonstrated their utility in rigid registration. Our representation addresses key limitations of discrete representations by enabling differentiable optimization and reducing computational overhead. More specifically, this approach trains a model on a given template mesh of arbitrary size, and subsequently replaces mesh-based interpolation with a simple forward pass through the pre-trained MLP.
Surface registration experiments on the ADRC dataset show that our approach achieves accuracy comparable to state-of-the-art methods while being consistently faster and more robust to large rotations. Integrating our optimization-based rigid initialization improves Dice scores and robustness in existing pipelines. Because the template NC model only needs to be fit once, its small pre-training cost is negligible, and using it for registration provides a clear runtime and computational advantage over classic mesh representations.

Future work will include evaluating our method on the MindBoggle dataset with another set of features to assess its robustness.
We also plan to integrate \textit{neural cortical maps} into an end-to-end neuroimaging pipeline with automatic cortical feature computation and NC-based NRR methods for more efficient and scalable surface-based analysis. 

\section{Compliance with ethical standards}
\label{sec:ethics}
This research study was conducted using human subject data made available in open access. 

ADNI dataset can be found at adni.loni.usc.edu, and ADRC dataset at people.csail.mit.edu/ythomas/code\_release.
\bibliographystyle{myIEEEbib}
\bibliography{refs}
\end{document}